# Improvised Salient Object Detection and Manipulation

Abhishek Maity [1]

[1] Department of Computer Science and Engineering, Guru Nanak Institute of Technology, India

*Abstract*— In case of salient subject recognition, computer algorithms have been heavily relied on scanning of images from top-left to bottom-right systematically and apply brute-force when attempting to locate objects of interest. Thus, the process turns out to be quite time consuming. Here a novel approach and a simple solution to the above problem is discussed. In this paper, we implement an approach to object manipulation and detection through segmentation map, which would help to de-saturate or, in other words, *wash out* the background of the image. Evaluation for the performance is carried out using the *Jaccard index* against the well-known *Ground-truth target box* technique.

## I. Introduction

Image salience or vision type saliency is the property through which we make some objects and items in the real world stand out from their surroundings so that it can grab our attention on first sight. With the advent of the digital camera, the quantity of visual data available on the internet is increasing exponentially. Thus, artistic photography of real-life images is becoming an important component in image computing and industry. Many well-known and sophisticated tools are available, but the cost associated with them may be quite high at few times. Using Image desaturation, Aesthetic imaging, Cartoon effects etc. are quite popular among people from all spheres. And most notably, many applications, like image display on miniature gadgets [22] etc. people want to display the specific area with the high interest factor.

A large number of interesting artistic changes and manipulations can be performed on images and video if the object can be sufficiently reliably detected, including operations like automatic refocusing, background blurring, exposure and white colour balance, and correct object search in the image. Here we are implementing and extending existing methods for detecting salient subjects in the image [1], [2], [3], [4]. Then use the result to show an interesting artistic change with manipulation, where desaturation of the background is executed and a simulated stereo view of the subject is generated, that can be viewable as an animated GIF [2].

## II. Literature Review

One of the toughest challenge in computer vision is the identification of the salient area of an image. Many applications (e.g., [5], [6], [7], and [8]) make use of these resources have led to many definitions and detection algorithms. Classically, saliency subject detection algorithms were generally developed for identifying the regions that a human eye would like to focus at the first sight. [9], [10], [11], [12], [13] and [14] Saliency of this type is essential for understanding human visual attention as well as for applications like the auto focusing. While some have concentrated on detecting a single main subject of an image set. [3], [15], [16] Thus, a lot has been carried out in this area which include several techniques and approaches. Some of the existing methods for salient visualization as mentioned in N. Bruce and J. Tsotsos [12], L. Itti and P. Baldi [17], L. Itti *et al.* [9], O.L.Meur *et al.* [14] and J. K. Tsotsos et al. [18] are based on the framework of bottom-up computation feature. Some studies [19] [8] [20] showed visual focusing helps in object tracking, recognition, and detection. Methods of T. Liu et al. [3] and S. Goferman et al. [2] are quite different from the above and developed great accuracy.

Apart from these, many approaches are based on application specific to that been developed till date. They include features like the context of the dominant subject being important as the salient objects themselves. Examples ranges from classification of images [23], summarized collection of images [24], video retargeting as mentioned in M. Rubinstein et al. [25], thumbnailing techniques [6] etc. The above mentioned approaches have worked well in finding salient objects and areas in all types of images, but there remained a scope for achieving better accuracy.

## III. Methodology

Here the entire process flow is outlined on a top level, and the evaluation metrics are also discussed. Detailed implementations is described in the forthcoming section. To detect the subject or salient object hereafter onwards, several saliency maps from the original image are computed first. The Saliency Map is a map arranged topographically that represents visual prominence of a corresponding visual scene. A saliency map in operation is a probability function that tells how likely or probably a pixel is a part of the salient object, using certain features of the given image as the criterion. Then, a segmentation map is generated that will optimize the objective function described in [3].

After this, images are restricted to one prominent object are taken into account, because in segmentation map if there are too many connected regions then, we select those that has reasonable region and has great variety in colour for appropriate computation. In the end, the segmentation map is used to perform desaturation operation and the 3D GIF animation is generated. Since the image sets appear with labeled rectangles that enclose the salient subject, the correctness of proposed algorithm is evaluated through comparison of the smallest box enclosing segmentation map to the ground-truth target box. Section IV shows the results of the above process.

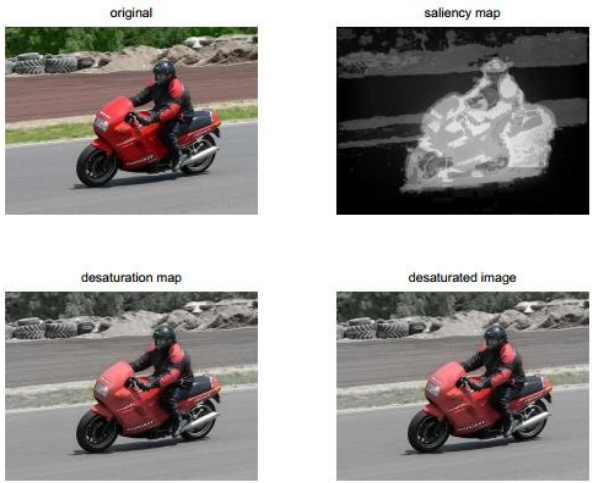

Fig. 1. The different images produced after computation

## IV. SALIENCY MAPS

In this part, the *seven* implementing saliency maps are presented in details, along with the algorithm used to optimize the main objective function. In Fig. 2, Fig. 3. and Fig. 4. entire processing flowchart is visualized in the form of a pipeline.

### A. Contrast on Multiple-Scale

In this paper, salient object are in focus is taken as a consideration for the saliency maps, and also "important" objects tend to maintain a lot of features of high contrast and edges. Thus, many a times reckoning the contrast gives a reasonable result of the salient object location. Through computing the multiple scale contrasting of the images, techniques that are quite robust in object size are established. The implementation procedure presented in [3] is followed closely for this. Results of the map obtained are given in Fig. 3a.

### B. Contrast on modified Multiple-Scale

The presented saliency map here is nothing but a modified multiple scale contrast map. As discussed earlier, we see that multiple-scale contrast as an effective detector for edge, but it is inefficient at identifying a region that is homogeneous in nature. Thus, for a simple example consisting of a black dot on a simple plain light background, multiple-scale map contrast correctly outlines the dot, but fails to realize the interior part of the dot is part is of the focused object.

To get the solution of this issue, a Revised or Modified form of Multiple-Scale Contrast map, as depicted in Fig. 3e. is introduced. The map takes the present multiple-scale map, local adaptive threshold is applied in the figure to identify all strong edges, and then removal of all regions completely surrounded by edges is initiated. Conversion of the present binary mask file into a probability map, in each connected region the highest probability in the corresponding area of the original Contrasting Multiple-Scale map is assigned. The reason of using the maximum instead of an average lies in the segmentation approach (discussed in subsection H.) used which is seen to be insensitive to values around 0.5, so using average factors will generate an inefficient map. As many different maps to perform segmentation is used, hence the use of extreme factors do not change the result to a great extent. Algorithm for multi-scale contrast s as follows.

Algorithm for *contrast on multi-scale*

1. Set *Gaussian Pyramid* level **L = 6.**
2. Convert image **RGB** to **Gray**.
3. Compute *imgsize = size(img)*
4. Apply **for** loop *l = 1:L* where
5.   *newimgsize = size(img)*
6.   *newcontrast = zeros(newimgsize)*
7.     Apply **for** loop *x = 2:newimgsize(2)-1*
8.     Apply **for** loop *y = 2:newimgsize(1)-1*
9.       Now, window=(img(y,x)-img(y-1:y+1,x1:x+1))^2
10.      *newcontrast(y,x) = sum(window(:))*
11.    End **innermost** loop
12.   End **inner** loop
13.   Now, return_image + = imresize(newcontrast, imgsize, *'bilinear'*)
14.   Apply img = imresize(img, 0.5, *'bilinear'*)
15. End **outer** loop
16. Return processed image map

### C. Content Based Saliency

This approach is adapted from [2], and is the only procedure for the existing code used. Optimizing an objective function through the factors of interesting object is as seen by human beings, including numerous psychological elements like the repeating patterns suppression, effect on regions with maximum contrast and the use of some beforehand available information. Fig. 3c. shows the things discussed through the map.

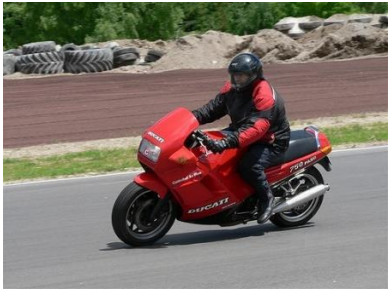

Fig. 2. Input Test Image

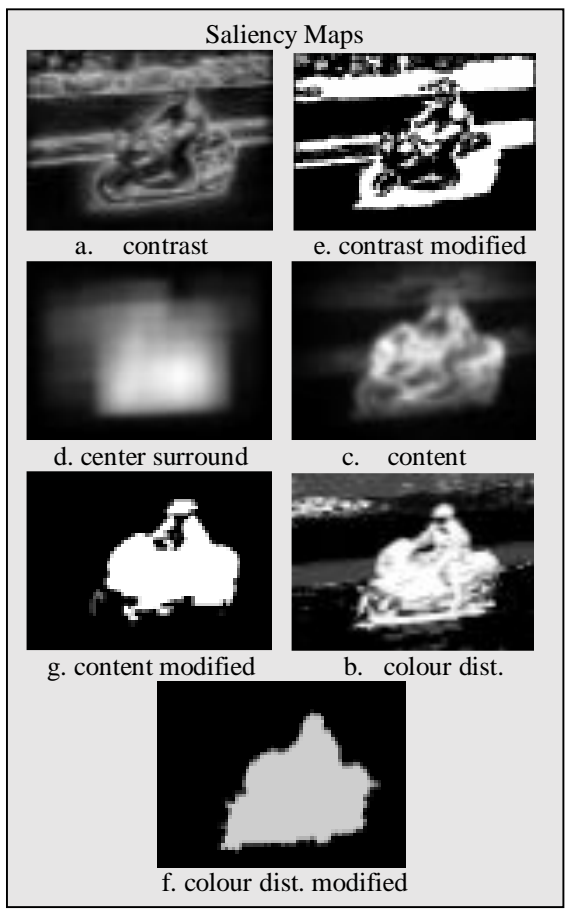

Fig. 3. Representation of various maps generated before desaturation. They are all processed to get final output.

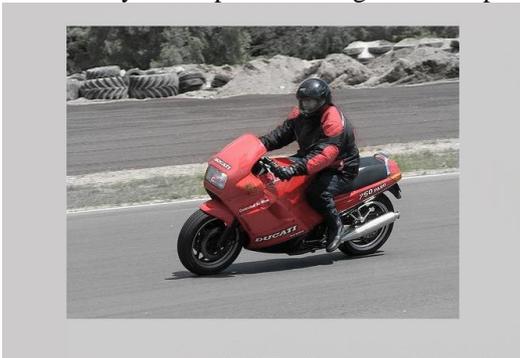

Fig. 4. Output image after processing.

*D. Modified Content Based Saliency*

The factual calculations shows that working of Content based Saliency is similar to that of an intelligent edge detector, where the edges belonging to the salient or prominent object is detected. Thus, processing described in subsection B, on Content Based Saliency map is executed to get the Modified Content Based Saliency map. Map of Fig. 3g. shows the result.

*E. Histogram centric surround saliency analysis*

The colours belonging to the salient subject is often quite different from its surroundings is an advantage to the salient map used here. Hence, if a box around the salient subject is drawn, and compared with the colour patterns inside the box to the colours surrounding the bounded area (in the part having the same area), there will be an expectation to view a large difference.

From the above observation, we can simply plot many different kinds of boxes and rectangles around a required pixel, and perform the computation of colour comparison. If the colours are not same to a great extent, then the selected pixel is most likely to be near the featured object. From analysis, reverse also holds. Now, if all the pixels are put on an iteration, an idea of probability of the salient object location is achieved. The in depth implementation can be obtained in [3]. However, executing exhaustive search for all pixels, the algorithm taking longer time space is observed. So, a down-sampling for the target image by a factor of 2 is made, and after execution up-sampling the solution from previous operation by a factor 2. This in turn solves the slow running of algorithm by factor of *four* with a minor accuracy loss. The findings of this map is shown in Fig. 3d. The algorithm is depicted below.

Algorithm for *centric surround analysis*

1. Compute center surround histogram **using**
2. **Apply** *if (nargin < 2)*
3.     **then** downsample = **4**
4. End *if* statement
5. **Set** *orginal_size = size(img)*
6. **Apply** *if (downsamp > 1)*
7.     **then** *img = imresize(img, 1/downsample)*
8. End *if* statement
9. **Set** *imgsize = size(img)*
10. **Set** *RECT_SIZE = [.1 .2 .3 .4 .5 .6] * min(imgsize(1:2))*
11. Generate an integral histogram
12. Find maximum chi-squared and rectangle properties at each pixel
13. Compute the center and surround histograms
14. Compute the center surround image
15.     **Check** *if (downsamp > 1)*
16. Return processed image map

*F. Spatial Distribution of Colour*

It is found that the saliency map used is the dual of Saliency Centric Histogram. While Saliency Centric Histogram looks for differences in local colour patterns, Spatial Distribution of Colour studies the behavior of each colour pattern that is distributed across the target image. If a particular colour spreads over a large area in the image, then it can concluded that the colour is not of that salient object. Whereas if a colour element is localized highly, there is a high probability that this colour is of the salient object. The sophisticated and complex mathematical calculations of colour distribution modeling is easily available in [3], and the results are used for implementation. Fig. 3c. highlights the inference of this map.

*G. Modified Spatial Distribution of Colour*

Since Spatial Distribution of Colour separate out colours, it is efficient at picking out chunks and parts instead of the edges, compared with Contrast Multiple-Scale and Content-Based Saliency. Since there will be only one salient subject in the target image, thus it is expected to have only one glob in the saliency map. With this conclusion, the Spatial Color Distribution map, performs global *Otsu* thresholding, remove all completely surrounded black regions, and then one connected part that is large in size and with rich colour content is selected. The volume of the convex hull feature surrounding any colour in RGB vector space as a metric for the color content is used. Lastly, the probability and chance of this connected area to the average connected region in the original saliency map is set. Fig. 3e. shows the discussed result of this map. Algorithm is illustrated and discussed below.

---

Algorithm for *spatial distribution of colour*

1. Shape image for distribution fit
2. Initial fit using *Gaussian mixture distribution*
3. Calculate pixel colour probabilities
4. Normalize probabilities **using**
5. img_prob=img_prob/**repma**t*(sum(img_prob,2), 1,numC)* **and**
6. img_prob=**reshape***(img_prob,[imgsize(1:2) numC])* **where** *numC = obj.NComponents*
7. Calculate variances
8. Calculate total normalized variance
9. Calculate variance-weighted sum
10. Return processed image map

---

*H. Generation of Segmentation Map*

To generate the segmentation map, Equation 3 in [3] is used as the objective function, reproduced here as:

$$E(A|I) = \sum_x \sum_{k=1}^{K} \lambda_k F_k(a_x, I) + \sum_{x,x'} S(a_x, a_{x'}, I) \quad (1)$$

$E(A|I)$ can be termed as "*energy*", a kind of measure penalty type, of the segmentation map $A$ for the given image $I$. The goal is to reduce and minimize $E(A|I)$ as far as possible. Here $x$ is one of the pixels of the processing image, whereas $x'$ is regarded as an adjacent pixel of $x$. The number of salient maps used is denoted by $K$. The weight assigned to the $k$th salient map is given by $\lambda_k$, and $F_k(a_x, I)$ is close to 0 if there is a match in segmentation map pixel and the salient map pixels, and value is nearly to 1 otherwise. $S(a_x, a_{x'}, I)$ function is the exponential decay of colour that is having low value when $x$ and $x'$ has the same label and colour, and maximum value otherwise.

On summarizing, the $\sum_x \sum_{k=1}^{K} \lambda_k F_k(a_x, I)$ measures how far off the segmentation map is from the salient maps. $\sum_{x,x'} S(a_x, a_{x'}, I)$ measures whether the assignment provides the label similar to adjacent color.

In [3], the salient map weights for each is evaluated based on machine learning from the image database. Thus, simplifying the process, roughly equal weights are assigned for each salient map, with some skews applied to favor maps that have shown to be effective greatly on images selected in a random fashion.

Now in order to generate the segmentation map, the map of first-order that optimizes only the first summation of the equation is computed (in other words, the adjacent pixels relationship of color, is ignored). This is achieved easily because of the absence of inter-pixel dependence. Then, the segmentation map's each pixel is checked, and values are inverted if there is an improvement in the objective function. This process is put into iteration until no such pixels exist, and we use the final result as the segmentation map.

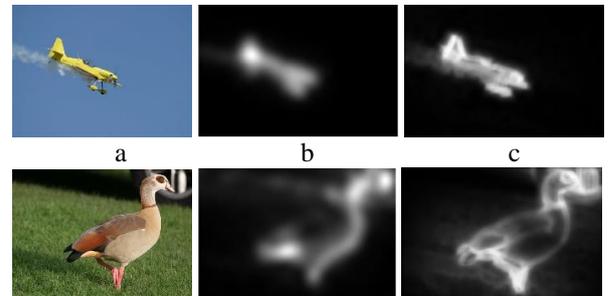

**a.** Input Images **b**. *Itti's* salient maps **c.** Our's salient maps

Figure 5a, 5b, 5c shows a comparison of the generated map with that of saliency map produced from *Itti's* algorithm [9]

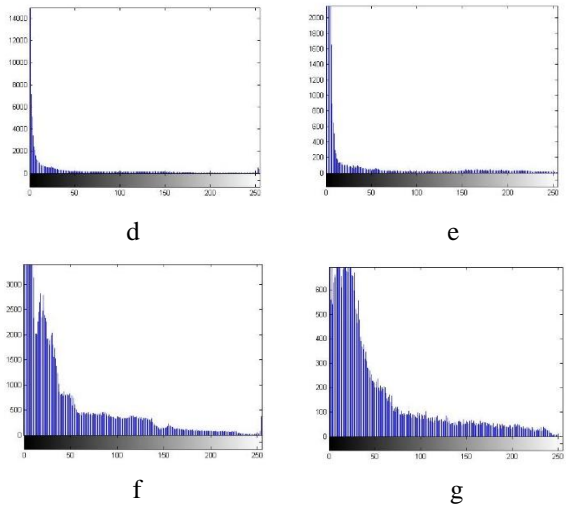

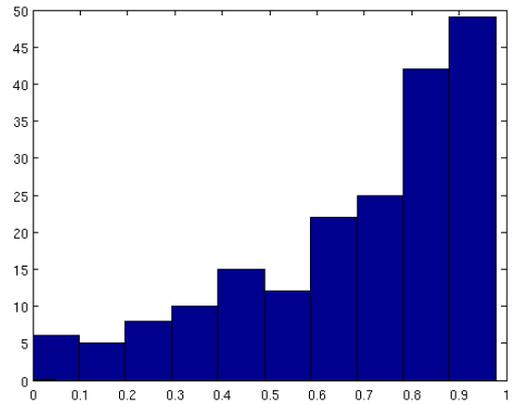

fig. 6. **Y-axis** denotes the *Histogram Count* and **X-axis** denotes the *Jaccard index*. Performance metric for the segmentation method by comparing against the ground truth target boxes using Jaccard index for 100 test data sets. The average Jaccard similarity index was 0.69.

Figure 5d and 5e are histogram representation of Itti's [9] and our's saliency map generated from second image respectively. Same is for 5f and 5g on second image respective to saliency maps.

## IV. RESULT AND ANALYSIS

A subset of 100 images out of the 5,000 from PASCAL VOC 2007 [21] database has been used for evaluation. In which each image has exactly one salient object or a group is taken as a unity, and for which the property of ground truth target bounding box of the salient subject is known or provided. The Jaccard similarity index is given by:

$$J(A,B) = \frac{|A \cap B|}{|A \cup B|} \quad (2)$$

With the Use of Jaccard similarity index of the bounding boxes property as the performance metric, it is found that accuracy increased from 58% to 69% over 100 images through the use of the new segmentation scheme and three derived saliency maps that are applied.

The entire computation was carried and tested using MATLAB R2013a (8.1.0.604) on 2.2GHz Intel(R) Core(TM) processor and x64 based architecture Personal Computer having a 4096 MB RAM. The image sets were obtained randomly from the mentioned database.

## V. CONCLUSION

Manipulation along with salient object detection is a tough image processing problem, and still is one of the popular topic of research. In this paper, study is carried on several recent papers with potential solution descriptions, and some algorithms are replicated and some of the modifications are added. It has been demonstrated that even without strong and robust machine learning application, satisfactory results with a reasonable number of images, 69% accuracy is achieved based on Jaccard's Index on a 100 sample images sets.

## VI. FUTURE WORK

Improvement that can be done significantly is the performance metric. As labels of only bounding boxes consisting the object were present in the ground truth data set, it was unable to use a more in-point specific criteria for performance, for example like object with a clean segmentation boundary. For example, the image of the sitting bird on the branch nearly maximizes our Jaccard index similarity criteria, but it's still quite away from obtaining a perfect segmentation. In fact, there appears to be some kind of systematic error in the size of the segmentation mask overestimating, an error which could better be accounted for using a better evaluation metric and precise labeled labeling set for ground truth. Then, more interesting changes and manipulation explorations can be done, including additional type saliency maps, along with deblurring that are depth-dependent. At last, accuracy of any of these mentioned methods can be improved to next level by using techniques of machine learning to build more and better classifiers for the saliency map data integration, for using the entire PASCAL VOC database of 5000 image sets for training and validation (compared with the 100).

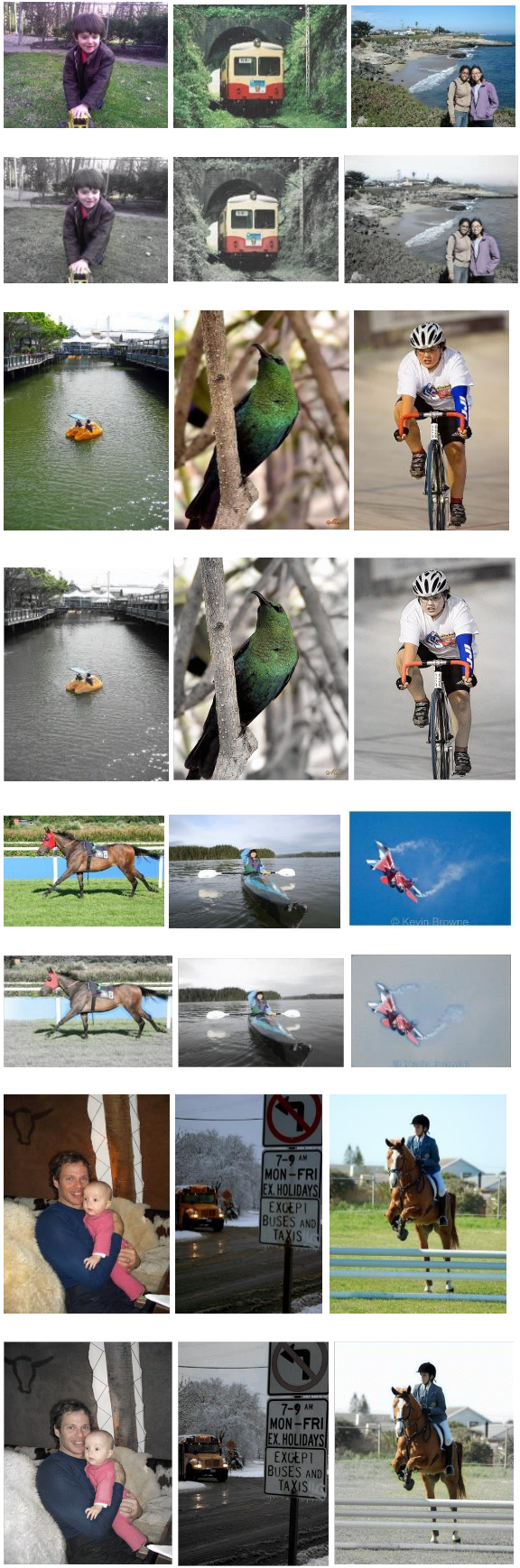

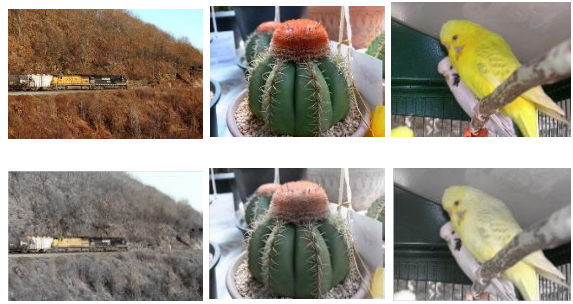

Fig. 7. Successful corresponding *top*-inputs *bottom*-outputs that have been desaturated using the segmentation map generated from the saliency maps.

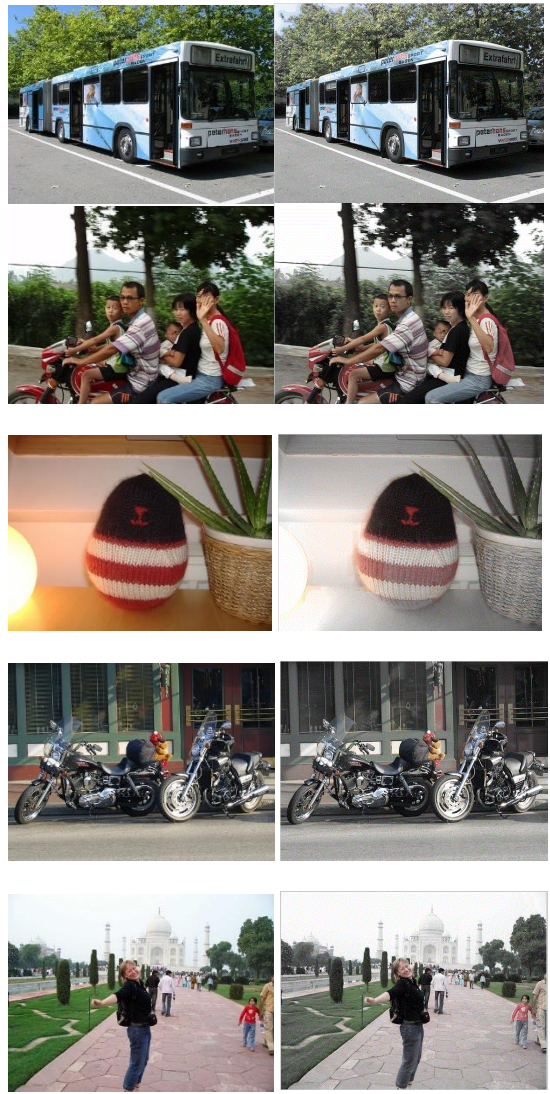

Fig. 8. Un-successful corresponding *top*-inputs *bottom*-outputs that have been desaturated using the segmentation map generated from the saliency maps.

Improvised Salient Object Detection and Manipulation

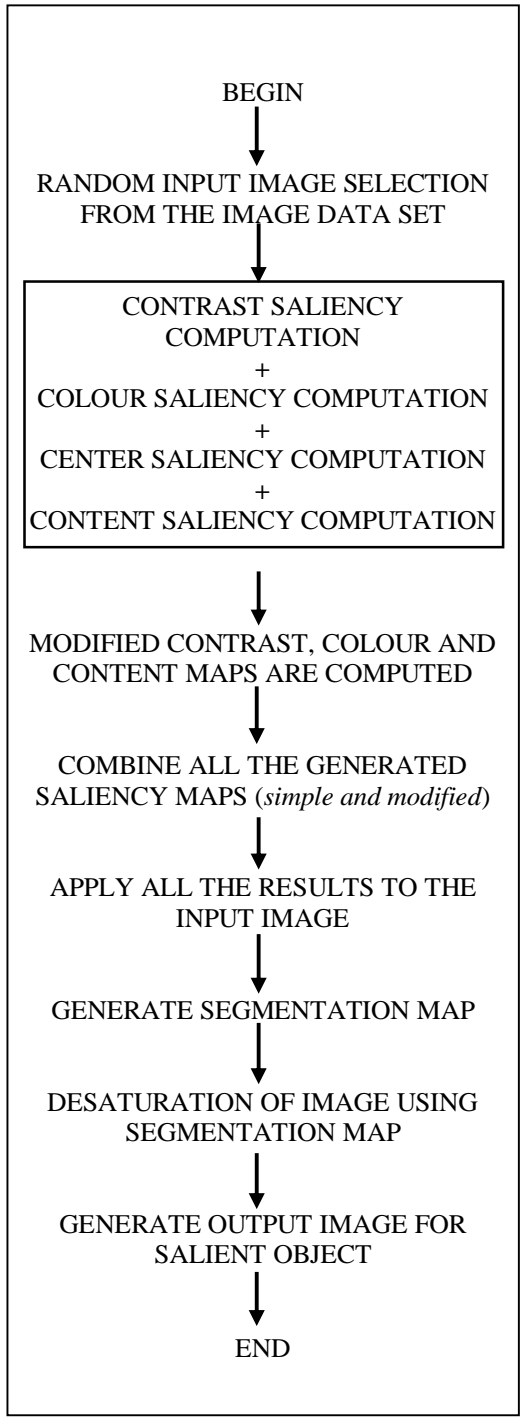

Fig. 9. Flow-chart for the entire process involved

ACKNOWLEDGMENT

The author would like to thank the anonymous reviewers' toward improvement of the paper.